\documentclass[journal]{IEEEtran}
\usepackage{lineno,hyperref}
\usepackage{bbm}
\usepackage{bm}
\usepackage{amsfonts}
\usepackage{amsmath}
\usepackage{mathrsfs}
\usepackage{amssymb}
\usepackage{enumerate}
\usepackage{graphicx}
\usepackage{subfigure}
\usepackage{booktabs}
\usepackage{graphicx}
\usepackage{setspace}
\usepackage{algorithm}
\usepackage{algorithmic}
\usepackage{setspace}
\usepackage{multirow}
\usepackage{array}
\usepackage{color}
\usepackage{cite}
\usepackage{geometry}
\usepackage{threeparttable} 
\usepackage{makecell}

\newcommand{\ie}{\textit{i}.\textit{e}.}
\newcommand{\eg}{\textit{e}.\textit{g}.}

\ifCLASSINFOpdf
\else
\fi
\begin{document}
\title{Language Prompt vs. Image Enhancement: Boosting Object Detection With CLIP in Hazy Environments}
\author{Jian~Pang, Bingfeng~Zhang, Jin~Wang, Baodi~Liu, Dapeng Tao, Weifeng~Liu,  
\thanks{J. Pang, B. Zhang, J. Wang, B Liu, W Liu, are with the College of Control Science and Engineering, China University of Petroleum (East China), Qingdao 266580, Shandong, P.R. China. (e-mail: jianpang@s.upc.edu.cn; bingfeng.zhang@upc.edu.cn; wangjin@s.upc.edu.cn; liubaodi@upc.edu.cn; liuwf@upc.edu.cn.). Corresponding author: Weifeng~Liu. }
\thanks{D. Tao is with the School of Information Science and Engineering, Yunnan University, Kunming 650091, Yunnan, P.R. China, and also with Yunnan United Vision Technology Co., Ltd., Kunming 650504, Yunnan, P.R. China. (e-mail:
dapeng.tao@gmail.com).}
}
\markboth{}%
{Shell \MakeLowercase{\textit{et al.}}: Bare Demo of IEEEtran.cls for IEEE Journals}
\maketitle
\begin{abstract}
Object detection in hazy environments is challenging because degraded objects are nearly invisible and their semantics are weakened by environmental noise, making it difficult for detectors to identify. Common approaches involve image enhancement to boost weakened semantics, but these methods are limited by the instability of enhanced modules. This paper proposes a novel solution by employing language prompts to enhance weakened semantics without image enhancement. Specifically, we design Approximation of Mutual Exclusion (AME) to provide credible weights for Cross-Entropy Loss, resulting in CLIP-guided Cross-Entropy Loss (CLIP-CE). The provided weights assess the semantic weakening of objects. Through the backpropagation of CLIP-CE, weakened semantics are enhanced, making degraded objects easier to detect. In addition, we present Fine-tuned AME (FAME) which adaptively fine-tunes the weight of AME based on the predicted confidence. The proposed FAME compensates for the imbalanced optimization in AME. Furthermore, we present HazyCOCO, a large-scale synthetic hazy dataset comprising 61258 images. Experimental results demonstrate that our method achieves state-of-the-art performance. The code and dataset will be released.
\end{abstract}
\begin{IEEEkeywords}
Object detection, CLIP, Approximation of Mutual Exclusion, HazyCOCO.
\end{IEEEkeywords}
\IEEEpeerreviewmaketitle
\section{Introduction}
\label{sec:intro}
Recently, deep learning models have achieved promising performance in object detection under clear and visible environments~\cite{Intro-1, Intro-2, Intro-3, Intro-4, Intro-5}. However, in visually degraded conditions such as hazy environments, their performance drops significantly. Object Detection in Hazy Environments (ODHE) is particularly challenging due to the prevalence of degraded objects, which are almost invisible and exhibit weakened semantic cues, making them extremely difficult to detect. For instance, in hazy scenes, persons or vehicles located deeper in the scene are more severely affected by environmental noise, resulting in blurred edges, reduced contrast, and corrupted textures. Consequently, detectors struggle to accurately classify and localize such degraded objects based on compromised semantics. This limitation poses a serious challenge for safety-critical applications such as autonomous driving~\cite{TCSVT-intro6}, intelligent surveillance~\cite{TCSVT-intro7,TCSVT-intro8}, and UAV navigation~\cite{TCSVT-intro9}, where reliable detection under adverse conditions is essential.

\begin{figure}
\centering
\includegraphics[width=3.3in, height=1.6in]{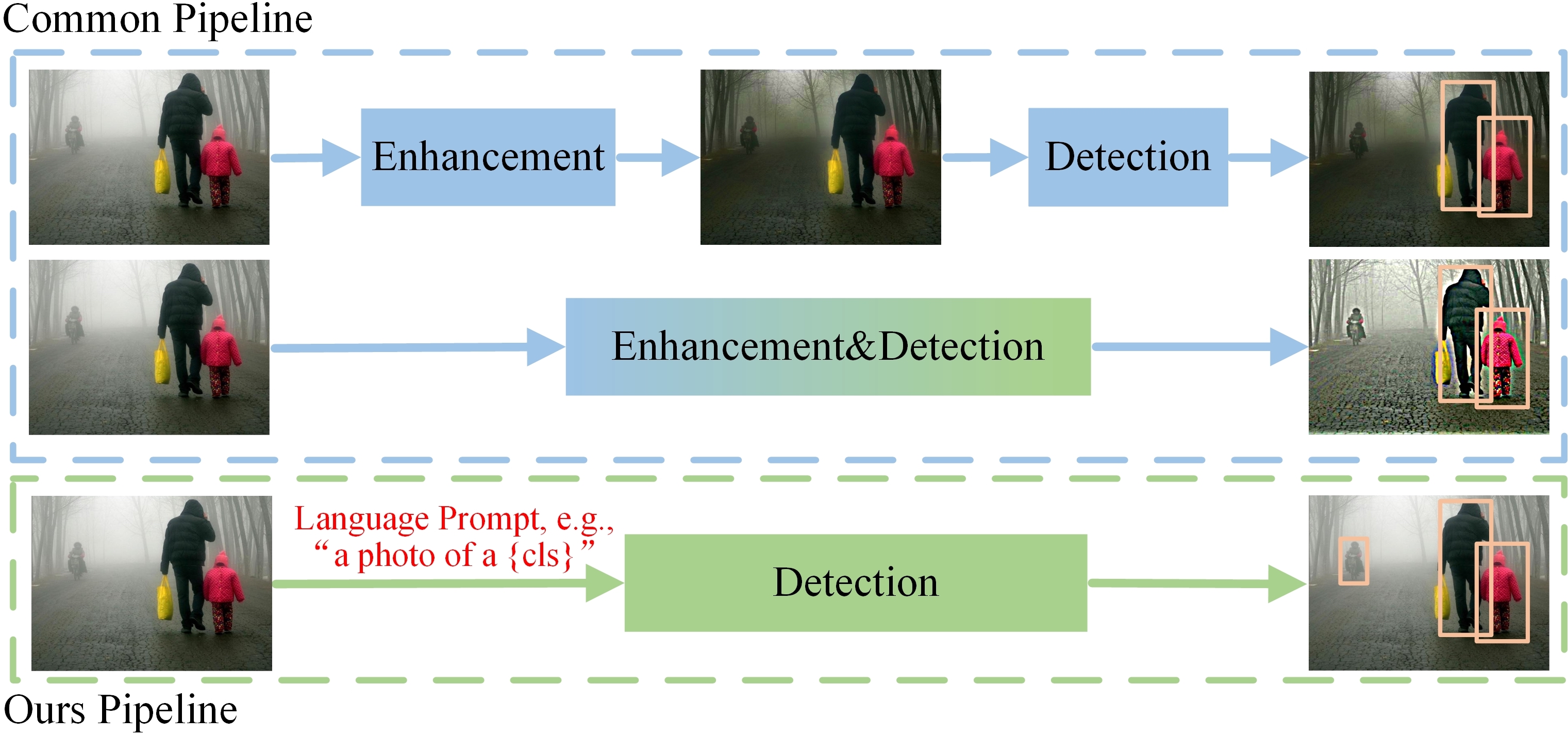}
\caption{Pipeline comparison. Our method can detect overlooked objects using only language prompts.}\label{fig: pip compare}
\end{figure}

\begin{figure}
\centering
\includegraphics[width=3.2in, height=1.9in]{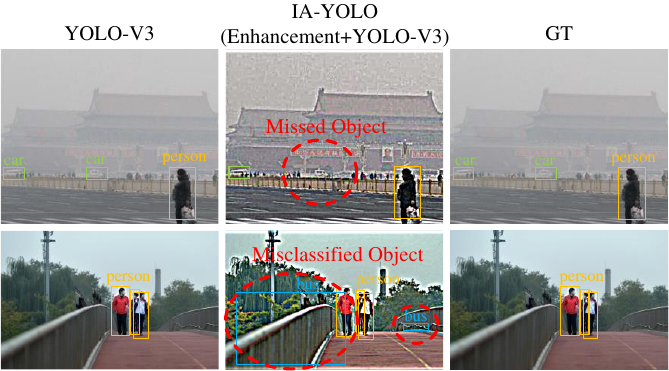}
\caption{Failure cases of IA-YOLO. GT: Ground Truth. Involving image enhancement may lead to missed detections or misclassifications.}\label{fig: failure}
\end{figure}

Currently, most common approaches employ image enhancement modules (\eg, image dehazing modules) to enhance weakened semantics~\cite{IAYOLO, friendly, Togethernet, advancing-dehazy}. As shown in Fig.~\ref{fig: pip compare}, the image enhancement module is linked to the detection module either in series or fusion. For methods with a series connection~\cite{advancing-dehazy}, the enhancement module and detection module are independent. Their objective is to obtain visually pleasant images for object detection. In contrast, methods with a fusion connection do not focus on recovering visually pleasing images. Instead, they optimize the enhancement module to generate detection-friendly semantic representations using detection loss~\cite{IAYOLO} or customized loss~\cite{friendly, Togethernet}. However, experiments reveal that the image enhancement module exhibits instability in two aspects. (1) The enhanced regions are uncontrollable: enhancement modules treat every pixel in the image equally, leading to inaccurate enhancement of degraded objects. For instance, if the background occupies most of the image, the enhancement may preferentially enhance the background and overlook the objects. (2) CNN-based enhancement modules can introduce human invisible noise~\cite{friendly}, which may result in misclassifications during object detection. Fig.~\ref{fig: failure} shows failure cases of IA-YOLO~\cite{IAYOLO} to verify our opinion. Based on the aforementioned analysis, we believe that image enhancement may not be a suitable choice for ODHE.

To compensate for weakened semantics, we involve CLIP which has been proven to provide credible semantic priors~\cite{PICLIP}. In this work, we present a CLIP-guided Cross-Entropy loss (CLIP-CE) to enhance the weakened semantics of degraded objects. Compared with CE, the CLIP-CE provides credible weights for degraded objects. The provided weight is calculated using the \textbf{A}pproximation of \textbf{M}utual \textbf{E}xclusion (AME) and \textbf{F}ine-tuned AME (FAME). In AME, we employ CLIP to compute similarities between the object and mutually exclusive prompts. These prompts consist of a positive prompt ``a photo of a \{\emph{cls}\}" and a negative prompt ``a photo without \{\emph{cls}\}". The negative similarity represents the degree of semantic weakening. For instance, if an object is heavily degraded, the negative similarity between the object and the negative prompt will be higher. We directly incorporate the negative similarity as the weight. With CLIP-CE, the model pays more attention to objects with weaker semantics. FAME is a fine-tuned version of AME. As training progresses, continuously assigning a larger weight to degraded objects leads to imbalanced optimization. Hence, we present FAME to balance the estimated weights in AME for further improvement. Specifically, we involve an adapter~\cite{adapter} after CLIP and rely on the predicted confidence of the object to fine-tune the weight. The principle behind FAME is to decrease the weights of objects with high predicted confidence and increase the weights of objects with low predicted confidence. Our method differs from Focal Loss (FL)~\cite{focalloss}, which is designed to focus on hard samples by assigning larger weights to those with lower predictions. In Section \ref{drawbacks of FL}, we analyze the principles of FL and explain why the weights estimated by FL are not suitable for ODHE.


Furthermore, we present the HazyCOCO dataset, a large-scale synthetic hazy dataset consisting of 61,258 images designed for ODHE. Compared to the commonly used datasets, such as VOC Fog~\cite{IAYOLO}, HazyCOCO has two advantages: (1) to the best of our knowledge, it is the largest synthetic hazy dataset for ODHE; (2) HazyCOCO better approximates real-world hazy conditions by using only outdoor scenes and reliable depth information. To generate reliable depths, we design a clamping operation on the relative depth, enhancing their hierarchical structure. This operation ensures that the generated hazy images closely mimic real-world hazy conditions. Our contributions are summarized as follows:
\begin{itemize}
    \item To the best of our knowledge, this is the first work to leverage a vision-language model to boost object detection in hazy environments.
    
    \item To enhance weakened semantics, we propose the CLIP-guided Cross-Entropy loss (CLIP-CE), which comprises the Approximation of Mutual Exclusion (AME) and Fine-tuned AME (FAME). CLIP-CE assigns credible weights to boost the weakened semantics of degraded objects.

    \item We introduce HazyCOCO, the largest synthetic hazy dataset for ODHE. Moreover, HazyCOCO ensures that the generated hazy images closely resemble the real world.

    \item The experimental results demonstrate that CLIP-CE outperforms existing approaches, achieving state-of-the-art (SOTA) performance in hazy environments. Moreover, CLIP-CE exhibits high generalizability, proving effective in real-world underwater and low-light conditions.
\end{itemize}

\begin{figure*}
\centering
\includegraphics[width=6.0in, height=3.3in]{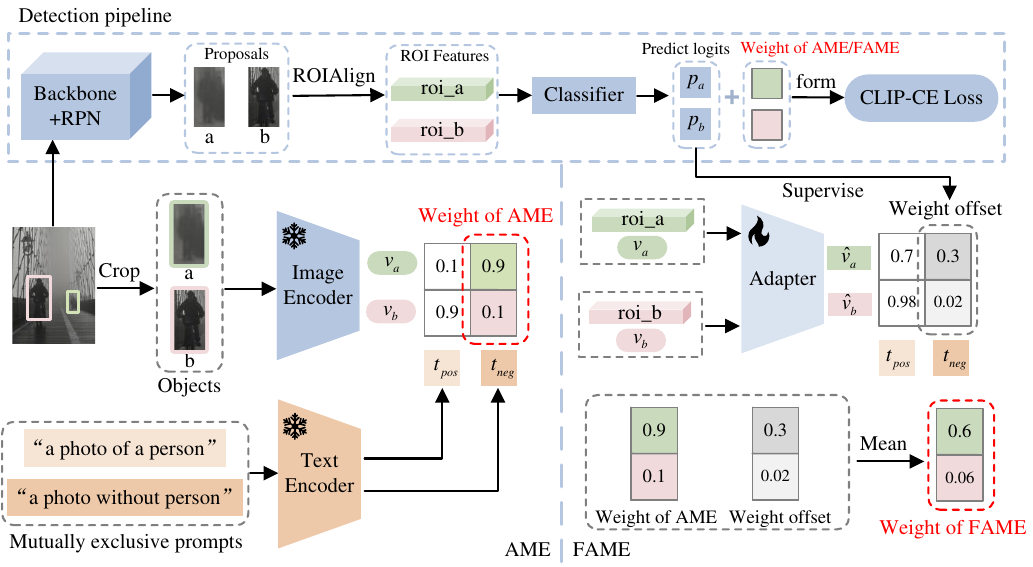}
\caption{Overall of our pipeline. \textbf{In detection pipeline:} The image is processed to obtain ROI features. The predicted logits of objects are combined with estimated weights to form CLIP-CE. \textbf{In AME:} We crop objects within the image and extract their visual embeddings. Meanwhile, we get text embeddings from mutually exclusive prompts. We calculate the negative similarity between these embeddings to determine the weight of AME. \textbf{In FAME:} We concatenate the ROI features with the visual embeddings to create adapted visual embeddings. We use negative similarities from adapted visual embeddings as weight offset. The final weight for FAME is averaged from the AME weight and the weight offset.
}
\label{fig: method}
\end{figure*}

The remainder of the paper is organized as follows. We first review the related works in Section \uppercase\expandafter{\romannumeral2}; Section \uppercase\expandafter{\romannumeral3} describes the proposed method in detail; Section \uppercase\expandafter{\romannumeral4} reports extensive experimental results and analysis; and finally the paper is summarized in Section \uppercase\expandafter{\romannumeral5}.

\section{Related Works}
\label{sec:rw}

\subsection{Object Detection in Hazy Environments} 
Object detection in hazy environments (ODHE) aims to locate and classify objects of interest in images captured under adverse conditions. In such environments, degraded objects (almost invisible) are widespread, and their semantics are weakened by environmental noise, leading to detection failures. Common approaches involve image enhancement to assist the detector~\cite{IAYOLO, friendly, DSNet, Togethernet, guided2023tnnls, Gdip}. These methods can be divided into two categories, (1) enhancement modules used to generate enhanced images for detection~\cite{IAYOLO, Gdip, friendly, Togethernet}. For instance, IA-YOLO~\cite{IAYOLO} employs a module that integrates traditional image processing methods to enhance the image, and the resulting enhanced image is input to YOLO-V3~\cite{yolov3} for detection. (2) Enhancement modules used to generate enhanced semantic representations~\cite{guided2023tnnls, DSNet, TCSVT-CFMW}. For example, IEG~\cite{guided2023tnnls} employs contrastive learning~\cite{contrastivelearning} between the image feature and the enhanced image feature to explore additional semantics. To the best of our knowledge, few studies have explored using vision-language models to boost ODHE. We believe that language prompts can compensate for the weakened visual semantics of degraded objects, thereby improving their visual representation.

\subsection{Vision-Language Models}
Vision-language models have been widely studied, numerous works utilize image-text pairs to learn a shared embedding space~\cite{CLIP, virtex, TCSVT-VLM2}. This space can be used for image classification, retrieval, and zero-shot prediction on unseen labels. CLIP~\cite{CLIP} collects 400 million image-text pairs from the Internet and uses a contrastive pre-training method to learn image-text relationships. CLIP enhances Open Vocabulary tasks, such as Open Vocabulary Semantic Segmentation and Open Vocabulary Detection, allowing for the detection or segmentation of objects using arbitrary text descriptions~\cite{RW1-AAAI, RW2-CVPR, TCSVT-VLM}. Recently, some approaches have employed CLIP to provide credible priors to support high-level vision tasks~\cite{PICLIP, VTCLIP}. For instance, PI-CLIP~\cite{PICLIP} uses CLIP with well-designed mutually exclusive prompts to generate credible masks for segmentation. Unlike these works, we propose a novel loss function that uses CLIP to provide credible weights for degraded objects. Through backpropagation of this loss, the weakened semantics are enhanced, thereby improving detection performance.

\section{Proposed Method}
\label{sec:met}

\subsection{Overview of Method}
As shown in Fig.~\ref{fig: method}, our method consists of three parts: the detection pipeline, Approximation of Mutual Exclusion (AME) and Fine-tuned AME (FAME). The detection pipeline employs Faster R-CNN~\cite{fasterrcnn} as experimental results indicate that recently introduced detectors, such as RT-DETR~\cite{rtdetr} and Deformable-DETR~\cite{deformable-detr}, do not demonstrate a significant performance advantage in hazy environments. more analysis can be referred to sec~\ref{sec:gen of clipce}.
We replace the Cross-Entropy loss (CE) used in Faster-RCNN with our CLIP-guided Cross-Entropy loss (CLIP-CE). In the following, we first introduce CE and its variants, analyzing their drawbacks when applied to ODHE. Our objective is to improve CE for ODHE. Therefore, we present AME and FAME which provide credible weights for CE and result in the development of CLIP-CE.

\subsection{Drawbacks in Cross-Entropy Loss and its Variants}
\label{drawbacks of FL}
In object detection, the Cross-Entropy loss (CE) plays a crucial role in model performance. The CE loss is defined as follows: 
\begin{equation}
\begin{aligned}
    {\mathcal{L}}_{\text{CE}}=-\sum_{c=1}^{C}{y}_{c}log\left ({p}_{c}\right )
\end{aligned},
\end{equation}
$C$ is the number of classes. ${p}_{c}$ is the predict probability for the $c$ class. ${y}_{c}$ is the $c$-th element in one-hot vector, and defined as follows:
\begin{equation}
\begin{aligned}
    {y}_{c}=\begin{cases} 
1& c=GT \\ 
0& otherwise
\end{cases}
\end{aligned},
\end{equation}
$GT$ is the ground truth. For notational convenience, we rewrite ${\mathcal{L}}_{\text{CE}}=-\sum_{c=1}^{C}{y}_{c}log\left ({p}_{c}\right )=-log\left({p}_{t}\right)$, ${p}_{t}$ is the predict probability for the ground truth. The CE treats each object with equal importance. However, in hazy environments, numerous degraded objects with weakened semantics require additional attention. Consequently, CE is not suitable for ODHE.

Focal Loss (FL)~\cite{focalloss} is an improved version of CE, it designs a novel weight for hard detected objects (predict probability is low):
\begin{equation}
\begin{aligned}
    {\mathcal{L}}_{\text{FL}}=-{\left( 1-{p}_{t}\right)}^{\gamma}log\left({p}_{t}\right)
\end{aligned},
\end{equation}
where a weight item ${\left( 1-{p}_{t}\right)}^{\gamma}$ is involved. This weight prompts FL to pay greater attention to objects with low predictions. However, our experiments reveal that during training, FL can easily lead to overfitting and subsequently decrease the weights assigned to degraded objects. For instance, FL initially assigns high weights to degraded objects. However, as training continues, these weights decrease significantly, leading to insufficient attention to the degraded objects. The detailed analysis can be found in experiments~\ref{exp: weight visual}.

\subsection{Approximation of Mutual Exclusion}
The objective of Approximation of Mutual Exclusion (AME) is to provide credible weights for degraded objects in CE, thereby making the improved CE suitable for ODHE. Given an image $I\in {\mathbb{R}}^{H\times W\times3}$, we crop objects within the image, forming a set $\left\{{b}_{i}\right\}_{i=1}^{m}$, where ${b}_{i}$ is the $i$-th object. For each object ${b}_{i}$, we define a pair of mutually exclusive texts ${e}_{i}^{\text{pos}}$ and ${e}_{i}^{\text{neg}}$. Here, ${e}_{i}^{\text{pos}}$ is a positive text, \eg, ``a photo of a \{\emph{cls}\}", ${e}_{i}^{\text{neg}}$ is a negative text, \eg, ``a photo without \{\emph{cls}\}", where \{\emph{cls}\} represents the class of ${b}_{i}$.

Let $\mathcal{V}$ and $\mathcal{T}$ denote the visual encoder and text encoder in CLIP, respectively. The visual embedding of $b_i$ is  ${v}_{i}=\mathcal{V}\left({b}_{i}\right)$. The mutually exclusive text embeddings of $b_i$ are ${t}_{i}^{\text{pos}}=\mathcal{T}\left({e}_{i}^{\text{pos}}\right)$ and ${t}_{i}^{\text{neg}}=\mathcal{T}\left({e}_{i}^{\text{neg}}\right)$. We measure the similarity between the visual embedding and the two mutually exclusive text embeddings as follows: $sim_{i}^{\text{pos}}=v_{i}^{\top} \times {t}_{i}^{\text{pos}}$ and $sim_{i}^{\text{neg}} =v_{i}^{\top} \times {t}_{i}^{\text{neg}}$. If the visual semantics of the object are weakened, the similarity $sim_{i}^{\text{neg}}$ will be higher than $sim_{i}^{\text{pos}}$. We apply softmax to approximate the score of negative similarity:
\begin{equation}
\label{Eq: AME}
\begin{aligned}
{w}_{i}^{\text{ame}}=\frac{e^{sim_{i}^{\text{neg}}}} {{e^{sim_{i}^{\text{neg}}} + e^{sim_{i}^{\text{pos}}}}}
\end{aligned},
\end{equation}
${w}_{i}^{\text{ame}}$ is the weight for object $b_i$, representing the degree of semantic weakening of $b_i$. For example, if an object is severely degraded and almost invisible in the image, the visual encoder in CLIP fails to capture its semantics, leading to $e^{sim_{i}^{\text{neg}}}$ higher than $e^{sim_{i}^{\text{pos}}}$, and ultimately resulting in a high ${w}_{i}^{\text{ame}}$. We incorporate ${w}_{i}^{\text{ame}}$ into the cross-entropy loss to form CLIP-CE (Eq.~\ref{Eq: CLIPCE}). With CLIP-CE, the detector pays greater attention to objects with weakened semantics.

\subsection{Fine-tuned Approximation of Mutual Exclusion}
In the previous section, we used mutually exclusive texts to approximate the weight for each object. However, since the CLIP model is frozen, these approximated weights remain constant. As training progresses, continuously assigning large weights to degraded objects leads to imbalanced optimization of the detector. To address this issue, we propose a Fine-tuned Approximation of Mutual Exclusion (FAME) to dynamically adjust the approximated weights in AME. Our objective is intuitive: as training progresses, high-confidence objects should no longer require large weights, while low-confidence objects should have their weights increased. To achieve this, we add an adapter~\cite{adapter} after the CLIP to fine-tune the approximated weights in AME.

In AME, the approximated weight depends solely on the visual embedding from CLIP, which is insufficient to determine whether the weight needs adjustment. Therefore, we incorporate both the object feature and the predicted probability from the detection pipeline for fine-tuning. The object feature is the basis for adjusting weight. The predicted probability guides the direction of the weight adjustment.

The object feature is obtained as follows: in the detection pipeline (Fig.~\ref{fig: method}), we first extract the feature map from the image. Then, we employ RoIAlign~\cite{fasterrcnn} based on positive proposals ($IoU > 0.5$) to obtain a set of object features $\left\{r_{i}^{h}\right\}_{h=1}^{H}$, where $H$ is the number of positive proposals. For simplicity, we assume $H = 1$. For the object $b_i$, its object feature $r_i$ and visual embedding $v_i$ are input to the adapter:
\begin{equation}
\begin{aligned}
\hat{v}_{i}=Adapter\left ( \left [v_i,r_i \right ] \right )
\end{aligned},
\end{equation}
where $\left[\cdot \right]$ is concatenation. $\hat{v}_{i}$ is the adapted visual embedding. Following AME, the similarities between the adapted visual embedding and the two mutually exclusive text embeddings are given by $\hat{sim}_{i}^{\text{pos}} = \hat{v}_{i}^{\top} \times t_{i}^{\text{pos}}$ and $\hat{sim}_{i}^{\text{neg}} = \hat{v}_{i}^{\top} \times t_{i}^{\text{neg}}$. Following Eq.~\ref{Eq: AME}, we approximate the score of the negative similarity as follows:
\begin{equation}
\begin{aligned}
{w}_{i}^{\text{\text{offset}}}=\frac{e^{\hat{sim}_{i}^{\text{neg}}}} {{e^{\hat{sim}_{i}^{\text{neg}}} + e^{\hat{sim}_{i}^{\text{pos}}}}}
\end{aligned},
\end{equation}
where ${w}_{i}^{\text{\text{offset}}}$ is the offset weight. We expect ${w}_{i}^{\text{\text{offset}}}$ to be dynamically adjusted according to the object feature. However, without supervision for the adapter, ${w}_{i}^{\text{\text{offset}}}$ will be ineffective. To address this, we use the predicted probability to build soft labels to supervise the adapter. The soft label is defined as follows:
\begin{equation}
\label{Eq: softlabel}
\begin{aligned}
    {u}_{i}=\begin{cases} 
0& p_{i,t} > \theta  \\ 
1& otherwise
\end{cases}
\end{aligned},
\end{equation}
where $p_{i,t}\in [0, 1]$ is the predicted probability for the ground truth of the $i$-th object, and $\theta$ is a probability threshold. We use binary cross-entropy loss to supervise the adapter as follows:
\begin{equation}
\label{Eq: Adapter}
\begin{aligned}
    {\mathcal{L}}_{\text{adapter}} = & -\left[ {u}_{i} \times \log\left({w}_{i}^{\text{offset}}\right) + \right. \\
    & \left.  \left(1 - {u}_{i}\right) \times \log\left(1 - {w}_{i}^{\text{offset}}\right) \right]
\end{aligned},
\end{equation}
with Eq.~\ref{Eq: Adapter}, ${w}_{i}^{\text{\text{offset}}}$ is dynamically adjusted according to the object feature from the detector. 

Here, we present FAME to compensate for the imbalanced optimization in AME. The FAME uses the offset weight $w_i^{\text{offset}}$ to fine-tune ${w}_{i}^{\text{ame}}$ as follows:
\begin{equation}
\begin{aligned}
   {w}_{i}^{\text{fame}} = \frac{{w}_{i}^{\text{ame}}+{w}_{i}^{\text{offset}}}{2}
\end{aligned}
\end{equation}

Finally, we present the proposed CLIP-guided Cross-Entropy loss (CLIP-CE):
\begin{equation}
\begin{aligned}
\label{Eq: CLIPCE}
    {\mathcal{L}}_{\text{CLIP-CE}}&=\begin{cases} 
-{e}^{\alpha_1 \times w_{i}^{\text{ame}}}\times log\left(p_{i,t}\right)& \text{if } \text{epoch} \in (0, ep_i] \\
-{e}^{\alpha_2 \times w_{i}^{\text{fame}}}\times log\left(p_{i,t}\right)& \text{if } \text{epoch} \in (ep_i, ep_j]
\end{cases}
\end{aligned}
\end{equation}
where $ep_i$ is the pre-training epoch and $ep_j$ is the total training epoch. For instance, during the training epoch within $\left ( 0, ep_i\right ]$, ${e}^{\alpha_1 \times w_{i}^{\text{ame}}}$ is used as the weight of the CE loss. Here, $p_{i,t}$ is the predicted probability for the ground truth of the $i$-th object.

\subsection{Training Objective}
Our training objective has two parts, one for object detection, and another for adapter. The objectives are as follows:
\begin{equation}
\begin{aligned}
    \mathcal{L}=\begin{cases} 
{\mathcal{L}}_{\text{rpn}}+{\mathcal{L}}_{\text{bbox}}+{\mathcal{L}}_{\text{CLIP-CE}},& \mathrm{for~detection}  \\ 
{\mathcal{L}}_{\text{adapter}},& \mathrm{for~adapter}
\end{cases}
\end{aligned}
\end{equation}

For object detection, we replace the cross-entropy loss in Faster-RCNN with our $\mathcal{L}_{\text{CLIP-CE}}$. For $\mathcal{L}_{\text{rpn}}$ and ${\mathcal{L}}_{\text{bbox}}$, please refer to~\cite{fasterrcnn}.

\section{Experiments}
\label{sec:exp}

\subsection{Datasets}
Our method is evaluated not only in real hazy environments, but also in real underwater and real low-light scenarios. The details of all datasets are shown in Tab.~\ref{tab: dataset}. 

\textbf{Low-light:} Exdark~\cite{Exdark} contains 7363 images collected from low light environment. We consider 5 classes in Exdark, \ie, person, car, bus, bicycle
and motorcycle.

\textbf{Underwater:} TrashCan~\cite{trashcan} has 7083 images which are observations of trash and a variety of undersea flora and fauna. We consider 18 classes, \eg, fish, crab, and bottle.

\textbf{Hazy environment:} We use RTTS~\cite{RTTS} and HazyCOCO for evaluation. RTTS is a real hazy dataset, we randomly divide it into 2161 images for training and the rest 2161 images for testing, the classes in RTTS are consistent with the Exdark. HazyCOCO is presented in this work, it is a large-scale synthetic hazy dataset, and has 80 classes which is consistent with the COCO2017~\cite{COCO}. 

\begin{table}[!t] \footnotesize
  \centering  
  \begin{threeparttable}    
    \begin{tabular}{ccccc}  
    \toprule  
    Dataset&Train&Test&Total&Environment\cr 
    \midrule  
    HazyCOCO&58800&2458&61258&Synthetic Haze\cr 
    RTTS&2161&2161&4,322&Real Haze\cr  
    Exdark&4411&2952&7363&Low light\cr  
    TrashCan&5936&1147&7083&Underwater\cr
    \bottomrule  
    \end{tabular}
    \caption{Image statistics of the datasets.}\label{tab: dataset}  
    \end{threeparttable}  
\end{table}


\begin{figure}
\centering
\includegraphics[width=3.2in, height=1.8in]{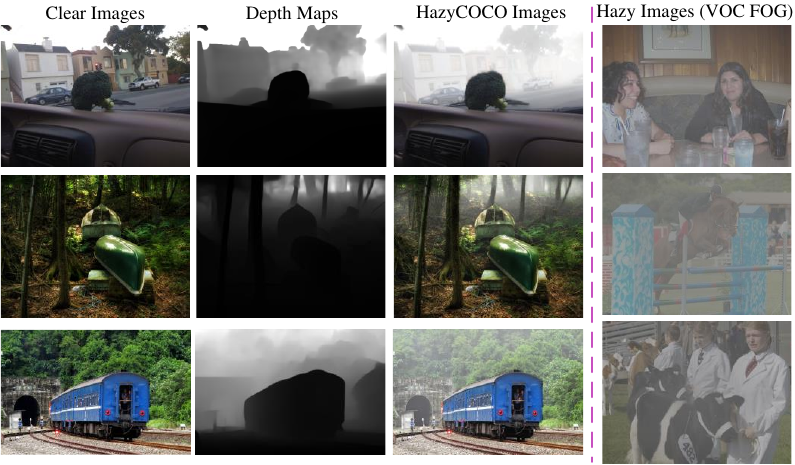}
\caption{Left part: Using the clear image and its depth map to generate HazyCOCO image. Right part: Hazy images from VOC FOG~\cite{IAYOLO}.}
\label{fig: hazycoco and vocfog}
\end{figure}

\subsection{HazyCOCO Generation}
\textbf{Problems in existing datasets:} (1) Insufficient data: The popular synthetic dataset VOC Fog~\cite{IAYOLO} only includes 10,845 images from Pascal VOC~\cite{VOC}. It is desired to build a large-scale dataset for object detection in hazy environments. (2) Deviate from real hazy images. The synthetic process of VOC Fog does not adhere to the atmospheric scattering model, resulting in unrealistic images. Furthermore, Pascal VOC includes many indoor images, synthesizing hazy images on these indoor images is inappropriate.

\subsubsection{Synthetic Details of HazyCOCO}
To address these issues, we synthesize the HazyCOCO dataset using the COCO2017 object detection dataset \cite{COCO}, which comprises 80 classes and is divided into train2017 (118,287 images) and val2017 (5,000 images). First, we employ Places365 \cite{places365} to filter out outdoor images from COCO2017, resulting in 58,800 images for training and 2,458 images for testing. Next, we use Depth-Anything~\cite{depthanything} to generate depth maps for each filtered outdoor image. Finally, we apply the Atmospheric Scattering Model (ASM)~\cite{atmospheremodel} to synthesize hazy images as follows:
\begin{equation}
\begin{aligned}
I(x)=J(x)t(x)+A(1-t(x))
\end{aligned},
\end{equation}
where $I(x)$ is the hazy image, $J(x)$ represents the hazy-free image. $A$ is the atmospheric light. $t(x)$ is medium transmission and can be expressed as:
\begin{equation}
\begin{aligned}
t(x)={e}^{-\beta d(x)}
\end{aligned},
\end{equation}
where $\beta$ is the scattering coefficient. $d(x)$ is the depth map corresponding to the hazy-free image. Given $J(x)$, $d(x)$, $\beta$ and $A$, the hazy image can be synthetic. Some generated images are given in Fig.~\ref{fig: hazycoco and vocfog}. In the following sections, we discuss how to obtain a reliable $d(x)$, as well as how to select appropriate values for $A$ and $\beta$.

\subsubsection{Clamp of Depth Map}
 In synthesis, the depth generated by Depth-Anything represents relative depth, which cannot be used directly. Since the relative depth shows a large discrepancy between the farthest and nearest pixels, caused the depth information at intermediate positions to be less discernible. To synthesize high-quality hazy images, one key factor is to provide the absolute depth map for ASM. However, in the COCO2017 dataset, it is not feasible to obtain the absolute depth, as the images are not captured using stereo or depth cameras. To tackle this, we use Depth-Anything~\cite{depthanything} to generate relative depth and design clamping operations to make the generated depth closer to real-world conditions.

\begin{figure*}[t]
\centering
\includegraphics[width=6.6in, height=2.2in]{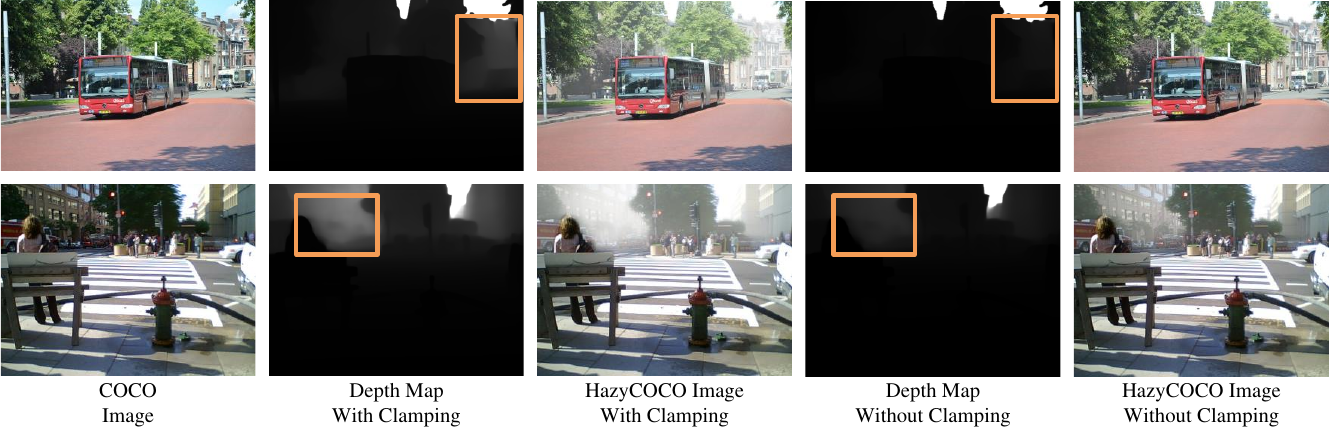}
\caption{Visualizations of the depth map with and without clamping. The depth map without clamping is directly generated by Depth-Anything. The HazyCOCO images are synthesized using the depth maps described above. In the depth map with clamping, the area within the orange box exhibits more hierarchical depth information.}
\label{fig: hazycoco_clamping}
\end{figure*}

In Fig~\ref{fig: hazycoco_clamping}, we show the limitations of the generated depth from Depth-Anything (denoted as ``depth without clamping"). The region within the orange box appears deep, but the depth map fails to accurately reflect the depth information. This issue arises because the relative depth shows a large discrepancy between the farthest and nearest pixels, causing the depth information at intermediate positions to be less discernible. To mitigate this, we clamp the farthest depth value to 100 times that of the nearest, thereby reducing the disparity between the farthest and nearest pixels. The clamped depth is shown in Fig.\ref{fig: hazycoco_clamping} (denoted as ``depth with clamping"). With the clamped depth, the generated hazy images exhibit a clearer sense of depth.

\subsubsection{Parameters Estimation}
\begin{figure*}[t]
\centering
\includegraphics[width=6.6in, height=2.2in]{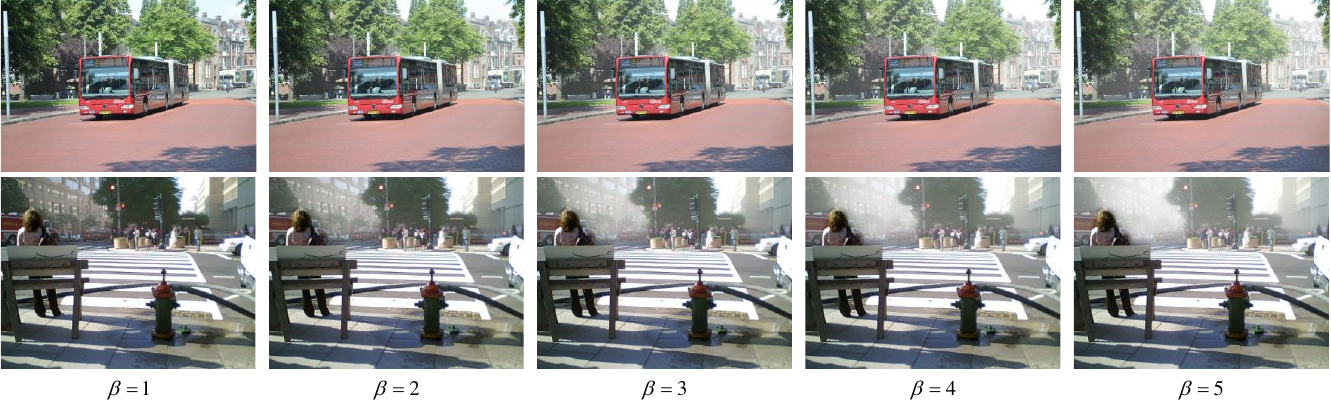}
\caption{Visualizations for the HazyCOCO images with different hazy densities.}
\label{fig: hazycoco_beta}
\end{figure*}

In image synthesis, $A$ controls the brightness, while $\beta$ controls the haze density. Directly setting $A$ as a constant may reduce image diversity; therefore, we employ DCP~\cite{DCP} to dynamically estimate $A$ for each image. To make the synthetic images more realistic, $\beta$ is randomly selected from the set $\left \{ 1,2,3,4,5 \right \} $. Fig.~\ref{fig: hazycoco_beta} shows synthetic images with different values of $\beta$, which cover a range of hazy densities commonly observed in real-world scenarios.

\subsection{Implementation Details}
\textbf{Model setting:} We utilize Faster-RCNN~\cite{fasterrcnn} for object detection, with a batch size of 4. We set the learning rate to 0.01 and employ stochastic gradient descent to optimize Faster-RCNN. The total training epoch is 20 ($ep_j=20$ in Eq.~\ref{Eq: CLIPCE}), and the pre-training epoch is 15 ($ep_i=15$ in Eq.~\ref{Eq: CLIPCE}). The CLIP~\cite{CLIP} uses the pre-trained ViT-B/32 model.  Our adapter comprises two fully connected layers, each followed by a ReLU~\cite{Relu}. The learning rate of the adapter is 0.01. Our experiments are conducted on two Tesla V100 GPUs and our code is based on the Pytorch. 

\textbf{Hyperparamters:} In Eq.~\ref{Eq: softlabel}, $\theta$ is 0.5 for all datasets. In Eq.~\ref{Eq: CLIPCE}, for HazyCOCO, $\alpha_1$ is 0.5 and  $\alpha_2$ is 1. For RTTS, Exdark, and TrashCan datasets, $\alpha_1=\alpha_2$, set to 2, 1, 0.5, respectively.

\textbf{Evaluation metrics:} We employ the mean Average Precision (mAP) as the evaluation metric and set IoU to 0.5 for all comparisons with other methods.

\subsection{Comparison with the State-of-the-Art}
We compared our method with state-of-the-art methods in hazy, low-light, and underwater environments. While, most existing methods focus on the hazy environment~\cite{Togethernet, DSNet, DRYOLO}, few consider two more adverse environments~\cite{IAYOLO}. To ensure a fair comparison, we evaluate these methods within their suitable environments. Note that, we reproduce all compared methods based on their released code, the parameters of each method follow the default setting in the paper. 

The compared methods are divided into three categories based on the training strategy: (1) Separation: the image is enhanced offline, and then used to train a detector. (2) Fusion: the image is used to train the image enhancement and the detector simultaneously. (3) Direct: the image is used directly to train the detector.

\textbf{Hazy environment:} Results are shown in Tab.~\ref{tab: hazy results}. In the separation strategy, Advancing~\cite{advancing-dehazy} and DCP~\cite{DCP} are dehazing algorithms. Advancing is based on neural networks, while DCP is a traditional method. The unsatisfactory results from Advancing may be attributed to the neural network introducing undesirable noise, which hinders detection performance. In contrast, DCP, which does not involve any neural network, achieves moderate performance. This suggests that neural network-based enhancements produce unstable images, thereby impairing detection. In the Fusion strategy, the performances of IA-YOLO~\cite{IAYOLO} and TogetherNet~\cite{Togethernet} are unpleasant, since these methods depend on the semantic quality of the enhanced image. However, in real hazy environments, enhancement algorithms often fail to restore rich semantic objects within the image. In direct strategy, we compared our method to Focal Loss~\cite{focalloss}, which involves customized weight into Cross-Entropy loss for exploring hard samples. Our method outperforms Focal Loss across two datasets.

\textbf{Low-light and underwater environments:} Results are shown in Tab.~\ref{tab: low-light results}.  In the separation strategy, RUSA~\cite{dark-ruas}, Zero-DCE~\cite{dark-ZeroDCE}, and UHDFour~\cite{dark-UHDFour} are low-light enhancement algorithms, while HCLR~\cite{HCLR} and UFormer~\cite{UFormer} are designed for underwater enhancement. In the direct category, BoostingRCNN~\cite{boostingRCNN} has recently been released for object detection in underwater environments. Based on the comparison, our method outperforms all the evaluated approaches, demonstrating its high generalizability and better performance.

\begin{table}[t] \footnotesize
  \centering  
  \begin{threeparttable}
    \begin{tabular}{>{\centering\arraybackslash}p{1.0cm}
                    >{\centering\arraybackslash}p{1.8cm}
                    >{\centering\arraybackslash}p{1.6cm}|
                    >{\centering\arraybackslash}p{1.2cm}
                    >{\centering\arraybackslash}p{0.8cm}} 
    \toprule  
    Strategy&Method&Detector&HazyCOCO&RTTS\cr
    \midrule
    \multirow{2}{*}{Separation}
    &Advancing~\cite{advancing-dehazy}&\multirow{2}{*}{Faster-RCNN}&36.77&70.29\cr
    &DCP~\cite{DCP}&&42.82&74.26\cr
    \midrule
    \multirow{2}{*}{Fusion}
    &IA-YOLO~\cite{IAYOLO}&YOLOV3&29.27&59.20\cr
    &TogetherNet~\cite{Togethernet}&YOLOX&32.31&54.00\cr
    \midrule
    \multirow{2}{*}{Direct}
    &Focal Loss~\cite{focalloss}&\multirow{2}{*}
    {Faster-RCNN}&39.67&70.20\cr
    &Ours&&\textbf{44.92}&\textbf{76.76}\cr
    \bottomrule  
    \end{tabular}  
    \caption{Performance comparison in hazy environment. Bold is the best. Results are reported by mAP.} 
    \label{tab: hazy results}
    \end{threeparttable}  
\end{table}

\begin{table}[t] \footnotesize
  \centering  
  \begin{threeparttable}
    \begin{tabular}{>{\centering\arraybackslash}p{1.0cm}
                    >{\centering\arraybackslash}p{1.8cm}
                    >{\centering\arraybackslash}p{1.6cm}|
                    >{\centering\arraybackslash}p{0.8cm}
                    >{\centering\arraybackslash}p{0.8cm}}   
    \toprule  
    Strategy&Method&Detector&Exdark&TrashCan\cr
    \midrule
    \multirow{5}{*}{Separation}
    &RUSA~\cite{dark-ruas}&\multirow{5}{*}{Faster-RCNN}&61.12&-\cr
    &Zero-DCE~\cite{dark-ZeroDCE}&&68.77&-\cr
    &UHDFour~\cite{dark-UHDFour}&&58.85&-\cr
    &HCLR~\cite{HCLR}&&-&48.23\cr
    &UFormer~\cite{UFormer}&&-&46.57\cr
    \midrule
    \multirow{2}{*}{Fusion}
    &IA-YOLO~\cite{IAYOLO}&YOLOV3&51.50&-\cr
    &TogetherNet~\cite{Togethernet}&YOLOX&-&44.22\cr
    \midrule
    \multirow{3}{*}{Direct}
    &BoostingRCNN~\cite{boostingRCNN}&\multirow{3}{*}{Faster-RCNN}&-&59.31\cr
    &Focal Loss~\cite{focalloss}&&64.50&46.96\cr
    &Ours&&\textbf{70.42}&\textbf{59.63}\cr
    \bottomrule  
    \end{tabular} 
    \caption{Performance comparison in low-light and underwater environments. Bold is the best. Results are reported by mAP. ``-" means that the method is not suitable for the environment.} 
    \label{tab: low-light results}
    \end{threeparttable}  
\end{table}

\begin{table}[t] \footnotesize
  \centering  
  \begin{threeparttable}
    \begin{tabular}{ccc} 
    \toprule  
    Method&Cross-Entropy Loss (CE)&CLIP-CE\cr
    \midrule
    Deformable-DETR~\cite{deformable-detr}&64.42&65.36\cr
    RT-DETR~\cite{rtdetr}&74.61&76.48\cr
    Faster-RCNN~\cite{fasterrcnn}&74.12&\bf{76.76}\cr
    \bottomrule
    \end{tabular}  
    \caption{Performance comparison of different detectors with and without the proposed CLIP-CE loss on the RTTS dataset.} 
    \end{threeparttable}  
\label{generalization}
\end{table}

\begin{table*}[ht] \footnotesize
  \centering  
  \begin{threeparttable}
    \begin{tabular}{ccccc}  
    \toprule  
    Methods&HazyCOCO (synthetic hazy)&RTTS (real hazy)&ExDark (real dark)&TrashCan (real underwater)\cr
    \midrule
    CE (baseline)&42.40 &74.12&68.59&52.03\cr
    CE w/ AME&44.58&75.91&69.77&58.52\cr
    CE w/ FAME&\textbf{44.92}&\textbf{76.76}&\textbf{70.42}&\textbf{59.63}\cr
    \bottomrule  
    \end{tabular}  
    \caption{Ablation study. CE: Cross-Entropy loss. AME: Approximation of Mutual Exclusion. FAME: Fine-tuned AME. ``CE w/ AME" means CE with the weight from AME, while ``CE w/ FAME" means CE weighted by FAME. Bold is the best. Results are reported by mAP.} 
    \label{tab: ablation}
    \end{threeparttable}  
\end{table*}

\begin{figure}[t]
\centering
\includegraphics[width=3.3in, height=2.3in]{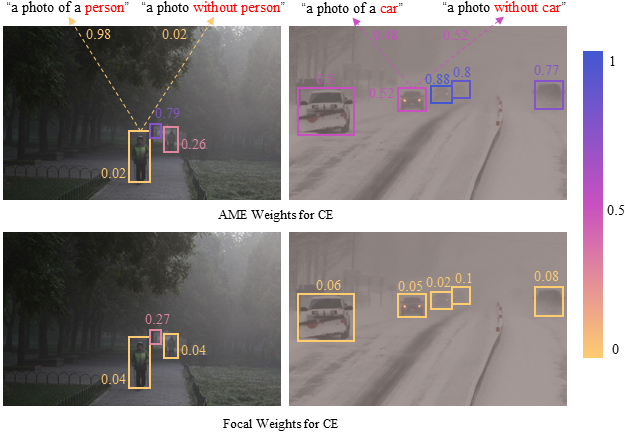}
\caption{Visualization for the AME weights and Focal weights. The AME weights are the similarity between object and negative prompt (``a photo without \{\emph{cls}\}"). The focal weight is computed from the predicted logit of object.}
\label{fig: weight visual}
\end{figure}

\subsection{Ablation Study}
\subsubsection{Effectiveness of the Proposed Components}
In this paper, we present the CLIP-guided Cross-Entropy loss (CLIP-CE). The CLIP-CE involves a novel weight item which estimated from AME or FAME. We evaluate the effectiveness of AME and FAME in common adverse environments. The results are shown in Tab.~\ref{tab: ablation}. All experiments in Tab.~\ref{tab: ablation} are based on the Faster-RCNN. Our Baseline uses the Cross-Entropy loss (CE). The CE with our AME weight (w/ AME) improved performance of the baseline on all datasets. Specifically, in the TrashCan dataset, the mAP is improved from 52.03\% to 58.52\%. It demonstrates the effectiveness of AME. Our Fine-tuned AME (w/ AME+FAME) further improved the performance based on the AME. It proves that the fine-tuned weight can compensate for the shortcomings of AME weight, thereby performance has further improvement.

\begin{figure*}
\centering
\includegraphics[width=6.6in, height=4.6in]{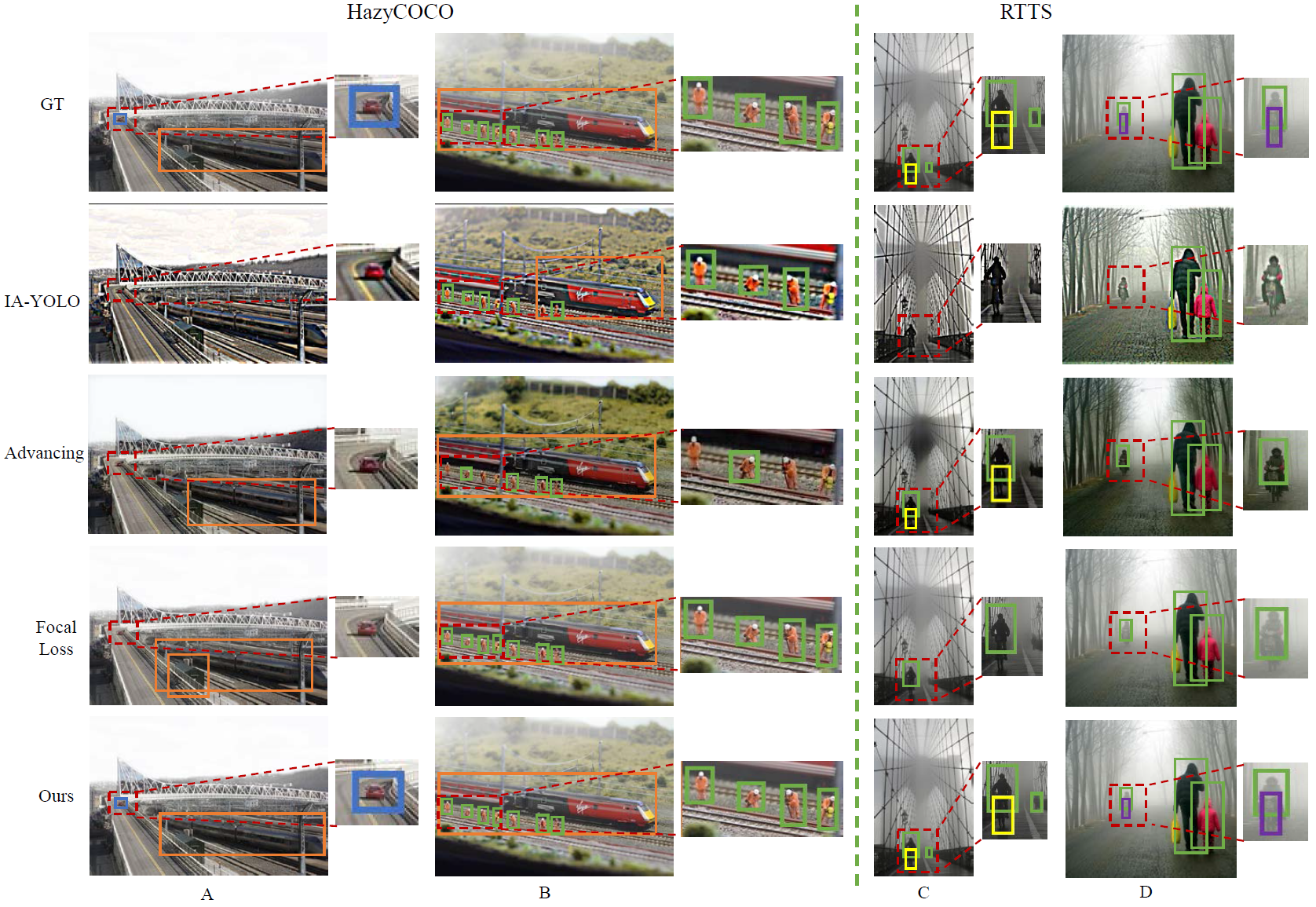}
\caption{Visualization of detection results on HazyCOCO and RTTS datasets. We recommend examining the zoomed-in areas (highlighted with red dashed boxes) for a clearer comparison. GT: ground truth.}
\label{fig: det results}
\end{figure*}

\begin{figure*}[t]
\centering
\includegraphics[width=6.2in, height=4.0in]{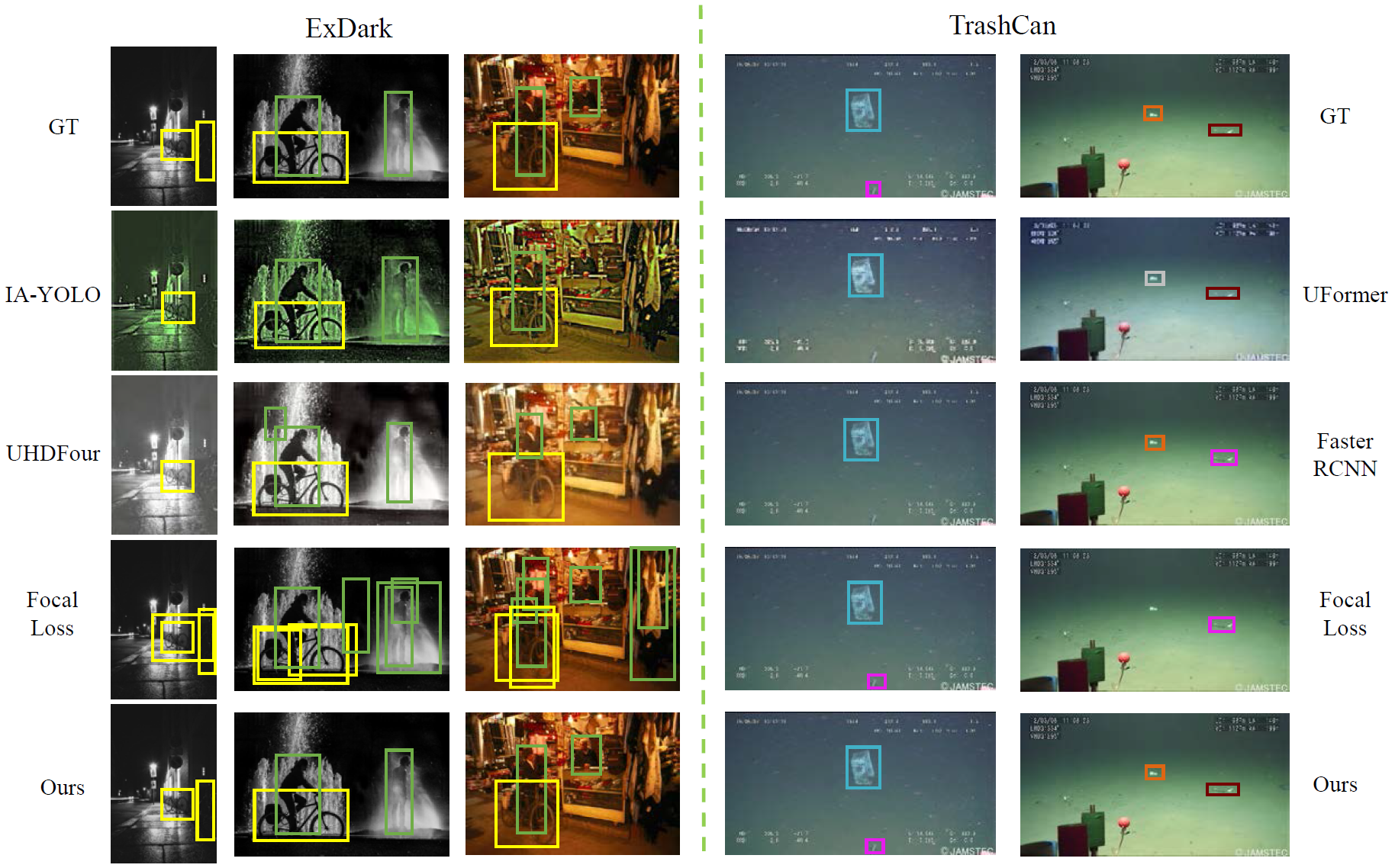}
\caption{Visualization of detection results. We recommend examining the zoomed-in areas for a clearer comparison. GT: ground truth. IA-YOLO, UHDFour and UFormer are image enhancement-based methods. Focal Loss refers to using Faster-RCNN with Focal loss. Ours is using Faster-RCNN with CLIP-CE.}
\label{fig: det results underwater_lowlight}
\end{figure*}

\subsubsection{Generalization to Other Detectors}
\label{sec:gen of clipce}
In tab.~\ref{generalization}, we embed our method into newly released detectors (Deformable-DETR~\cite{deformable-detr} and RT-DETR~\cite{rtdetr}), as well as a classical detector Faster R-CNN~\cite{fasterrcnn}. When integrated with our CLIP-CE loss, Deformable-DETR achieves an improvement from 64.42\% to 65.36\%, RT-DETR improves from 74.61\% to 76.48\%, and Faster R-CNN improves from 74.12\% to 76.76\% on the RTTS dataset. These consistent gains across different architectures demonstrate the strong generalization capability and versatility of the proposed method.


\subsection{Weight Visualizations and Analysis}
\label{exp: weight visual}
We visualize the AME weight and compare it with the Focal weight (as defined in Focal Loss~\cite{focalloss}). The AME weight represents the similarity between the object and the negative prompt (``a photo without \{\emph{cls}\}"). The Focal weight is computed as follows: $w={\left ( 1-p\right )}^{2}$, where $p$ is the predicted logit of an object. For instance, for an object, the lower the predicted confidence, the larger the Focal weight. The comparison is given in Fig.~\ref{fig: weight visual}. We observe that as the object becomes more degraded, the AME weight increases. For example, in the first columns, the weights for the three objects, from nearest to farthest, are 0.02, 0.26, and 0.79, respectively. In contrast, the Focal weights are not hierarchical, with the weights for the nearest, middle, and farthest objects being 0.04, 0.04, and 0.27, respectively. This issue may stem from overfitting caused by the use of Focal Loss, which leads the model to focus too heavily on degraded objects within the training set, resulting in high confidence for these objects. In contrast, our AME weight utilizes CLIP to provide reliable weights and mitigate overfitting.

\subsection{Prompt Comparison}
We evaluate the impact of different prompts on the performance. The experiments are conducted on the real hazy dataset RTTS. We select three types of prompts for comparison. Among the chosen prompts, ``a photo of a \{\emph{cls}\}” is used in our paper, we additionally add an environmental factor to it and raise a new prompt ``a foggy photo of a \{\emph{cls}\}”. Furthermore, we considered the prompt ``it is a \{\emph{cls}\}" for comparison, which carries a similar meaning to our prompt. The experimental results are shown in Tab.~\ref{tab: prompt}. The prompt ``a photo of a \{\emph{cls}\}" yielded the best results. While, the prompt ``a foggy photo of a \{\emph{cls}\}” with the environment factor demonstrates suboptimal performance, possibly because the environmental context diminished the significance of the class information. Similarly, the prompt ``it is a \{\emph{cls}\}", although semantically similar to our prompt, produced unsatisfactory results, which might be attributed to its infrequent usage in the CLIP. Our prompt, consistent with those commonly used prompts in CLIP, provides robust text-visual alignment, thereby contributing to its superior performance.

\begin{table}[!t] \footnotesize
  \centering  
  \begin{threeparttable}
    \begin{tabular}{>{\centering\arraybackslash}p{4.0cm}
                    >{\centering\arraybackslash}p{1.cm}
                    >{\centering\arraybackslash}p{1.cm}}   
    \toprule  
    \multirow{2}{*}[-2.5pt]{Prompt}&\multicolumn{2}{c}{RTTS}\cr
    \cmidrule(lr){2-3}
    &AME&FAME\cr
    \midrule
    ``a photo of a \{\emph{cls}\}” &\multirow{2}{*}{\bf{75.91}}&\multirow{2}{*}{\bf{76.76}} \cr
    ``a photo without \{\emph{cls}\}” &&  \cr
    \hline
    ``a foggy photo of a \{\emph{cls}\}” &\multirow{2}{*}{75.22}&\multirow{2}{*}{76.30} \cr
    ``a foggy photo without \{\emph{cls}\}” &&  \cr
    \hline
    ``it is a \{\emph{cls}\}” &\multirow{2}{*}{75.39}&\multirow{2}{*}{75.48} \cr
    ``it is not a \{\emph{cls}\}” && \cr
    \bottomrule  
    \end{tabular}  
    \caption{Prompt comparison on the real hazy dataset RTTS. The first prompt is used in our paper. The second prompt additionally has environmental factors. The third prompt is similar to ours. Bold is the best.} 
    \label{tab: prompt}
    \end{threeparttable}  
\end{table}

\subsection{Detection Visualizations}
\subsubsection{Visualizations for Hazy Environments}
To intuitively demonstrate the superiority of our method, we provide visualizations of the detection results in Fig.~\ref{fig: det results}, comparing our approach with IA-YOLO~\cite{IAYOLO}, Advancing~\cite{advancing-dehazy}, and Focal Loss~\cite{focalloss}. Both IA-YOLO and Advancing incorporate enhancement modules and apply object detection to enhanced images. Focal Loss means using the Faster-RCNN with Focal loss~\cite{focalloss}. In Fig.~\ref{fig: det results}, our method effectively detects objects that are overlooked by competitors, including those that are nearly unnoticeable to humans. For example, in the last column of Fig.~\ref{fig: det results}, the motorbike goes undetected by all compared methods. Although IA-YOLO produces enhanced images with richer textures and clearer colors, the detector still fails to identify the object. This reveals that visually appealing images may not contribute to object detection. Conversely, we believe that the key to success in this task lies in enhancing the semantic information of the objects, which may not depend solely on image quality. 
\subsubsection{Visualizations for Low-light and Underwater Environments}
We present detection results in low-light and underwater environments in Fig.~\ref{fig: det results underwater_lowlight}. In the low-light environment, the compared methods include IA-YOLO\cite{IAYOLO}, UHDFour~\cite{dark-UHDFour}, and Focal Loss~\cite{focalloss}. IA-YOLO integrates image enhancement with object detection, UHDFour~\cite{dark-UHDFour} is a recently introduced method for low-light enhancement, and Focal Loss refers to the use of Faster-RCNN with Focal Loss~\cite{focalloss}. In Fig.\ref{fig: det results underwater_lowlight}, our method effectively detects objects that are overlooked by competitors, including those that are nearly invisible to humans. In the underwater environment, UFormer\cite{UFormer} is a newly released method for underwater enhancement. Our comparison shows that while UFormer brightens the images, it fails to detect objects effectively. This highlights that visually appealing images may not necessarily improve object detection. Overall, our method outperforms existing approaches in common adverse environments, demonstrating its high generalizability.

\section{Conclusion}
In this paper, we present a novel solution for Object Detection in Hazy Environments (ODHE) by employing language prompts to enhance the weakened semantics of objects, rather than relying on image enhancement. The introduced CLIP-guided Cross-Entropy loss is highly generalizable and effective across common adverse conditions, such as hazy, low-light, and underwater environments. Compared to existing methods, our CLIP-CE demonstrates superior performance. The ablation study highlights the effectiveness of the proposed AME and FAME. Visualization of the detection results provides an intuitive comparison, revealing that our method can successfully detect objects overlooked by other approaches. The weight comparison shows that our weights are more suitable for ODHE. Furthermore, this work introduces HazyCOCO, the largest synthetic hazy dataset for ODHE, which ensures that the generated hazy images closely resemble real-world conditions.


%





\ifCLASSOPTIONcaptionsoff
  \newpage
\fi



%
\bibliography{mybibfile}

@String(CVPR= {IEEE Conf. Comput. Vis. Pattern Recog.})

@String(ICLR = {Int. Conf. Learn. Represent.})

@String(AAAI = {AAAI})

@String(CVPR  = {CVPR})

@String(ICLR  = {ICLR})

@article{atmospheremodel,
  title={Optics of the atmosphere: scattering by molecules and particles},
  author={McCartney, Earl J},
  journal={New York},
  year={1976}
}

@article{friendly,
  title={Detection-friendly dehazing: Object detection in real-world hazy scenes},
  author={Li, Chengyang and Zhou, Heng and Liu, Yang and Yang, Caidong and Xie, Yongqiang and Li, Zhongbo and Zhu, Liping},
  journal={IEEE Transactions on Pattern Analysis and Machine Intelligence},
  year={2023},
  publisher={IEEE}
}

@article{DSNet,
  title={DSNet: Joint semantic learning for object detection in inclement weather conditions},
  author={Huang, Shih-Chia and Le, Trung-Hieu and Jaw, Da-Wei},
  journal={IEEE transactions on pattern analysis and machine intelligence},
  volume={43},
  number={8},
  pages={2623--2633},
  year={2020},
  publisher={IEEE}
}

@inproceedings{IAYOLO,
  title={Image-adaptive YOLO for object detection in adverse weather conditions},
  author={Liu, Wenyu and Ren, Gaofeng and Yu, Runsheng and Guo, Shi and Zhu, Jianke and Zhang, Lei},
  booktitle={Proceedings of the AAAI Conference on Artificial Intelligence},
  volume={36},
  number={2},
  pages={1792--1800},
  year={2022}
}

@article{fasterrcnn,
  title={Faster r-cnn: Towards real-time object detection with region proposal networks},
  author={Ren, Shaoqing and He, Kaiming and Girshick, Ross and Sun, Jian},
  journal={Advances in Neural Information Processing Systems},
  volume={28},
  year={2015}
}

@inproceedings{Togethernet,
  title={TogetherNet: Bridging Image Restoration and Object Detection Together via Dynamic Enhancement Learning},
  author={Wang, Yongzhen and Yan, Xuefeng and Zhang, Kaiwen and Gong, Lina and Xie, Haoran and Wang, Fu Lee and Wei, Mingqiang},
  booktitle={Computer Graphics Forum},
  volume={41},
  number={7},
  pages={465--476},
  year={2022},
  organization={Wiley Online Library}
}

@article{RTTS,
  title={Benchmarking single-image dehazing and beyond},
  author={Li, Boyi and Ren, Wenqi and Fu, Dengpan and Tao, Dacheng and Feng, Dan and Zeng, Wenjun and Wang, Zhangyang},
  journal={IEEE Transactions on Image Processing},
  volume={28},
  number={1},
  pages={492--505},
  year={2018},
  publisher={IEEE}
}

@inproceedings{focalloss,
  title={Focal loss for dense object detection},
  author={Lin, Tsung-Yi and Goyal, Priya and Girshick, Ross and He, Kaiming and Doll{\'a}r, Piotr},
  booktitle={Proceedings of the IEEE international conference on computer vision},
  pages={2980--2988},
  year={2017}
}

@inproceedings{Gdip,
  title={Gdip: Gated differentiable image processing for object detection in adverse conditions},
  author={Kalwar, Sanket and Patel, Dhruv and Aanegola, Aakash and Konda, Krishna Reddy and Garg, Sourav and Krishna, K Madhava},
  booktitle={2023 IEEE International Conference on Robotics and Automation (ICRA)},
  pages={7083--7089},
  year={2023},
  organization={IEEE}
}

@article{guided2023tnnls,
  title={Image enhancement guided object detection in visually degraded scenes},
  author={Liu, Hongmin and Jin, Fan and Zeng, Hui and Pu, Huayan and Fan, Bin},
  journal={IEEE transactions on neural networks and learning systems},
  year={2023},
  publisher={IEEE}
}

@inproceedings{contrastivelearning,
  title={A simple framework for contrastive learning of visual representations},
  author={Chen, Ting and Kornblith, Simon and Norouzi, Mohammad and Hinton, Geoffrey},
  booktitle={International conference on machine learning},
  pages={1597--1607},
  year={2020},
  organization={PMLR}
}

@article{yolov3,
  title={Yolov3: An incremental improvement},
  author={Redmon, Joseph and Farhadi, Ali},
  journal={arXiv preprint arXiv:1804.02767},
  year={2018}
}

@inproceedings{virtex,
  title={Virtex: Learning visual representations from textual annotations},
  author={Desai, Karan and Johnson, Justin},
  booktitle={Proceedings of the IEEE/CVF conference on computer vision and pattern recognition},
  pages={11162--11173},
  year={2021}
}

@inproceedings{CLIP,
  title={Learning transferable visual models from natural language supervision},
  author={Radford, Alec and Kim, Jong Wook and Hallacy, Chris and Ramesh, Aditya and Goh, Gabriel and Agarwal, Sandhini and Sastry, Girish and Askell, Amanda and Mishkin, Pamela and Clark, Jack and others},
  booktitle={International conference on machine learning},
  pages={8748--8763},
  year={2021},
  organization={PMLR}
}

@inproceedings{depthanything,
  title={Depth anything: Unleashing the power of large-scale unlabeled data},
  author={Yang, Lihe and Kang, Bingyi and Huang, Zilong and Xu, Xiaogang and Feng, Jiashi and Zhao, Hengshuang},
  booktitle={Proceedings of the IEEE/CVF Conference on Computer Vision and Pattern Recognition},
  pages={10371--10381},
  year={2024}
}

@inproceedings{COCO,
  title={Microsoft coco: Common objects in context},
  author={Lin, Tsung-Yi and Maire, Michael and Belongie, Serge and Hays, James and Perona, Pietro and Ramanan, Deva and Doll{\'a}r, Piotr and Zitnick, C Lawrence},
  booktitle={Computer Vision--ECCV 2014: 13th European Conference, Zurich, Switzerland, September 6-12, 2014, Proceedings, Part V 13},
  pages={740--755},
  year={2014},
  organization={Springer}
}

@article{places365,
   title={Places: A 10 million Image Database for Scene Recognition},
   author={Zhou, Bolei and Lapedriza, Agata and Khosla, Aditya and Oliva, Aude and Torralba, Antonio},
   journal={IEEE Transactions on Pattern Analysis and Machine Intelligence},
   year={2017},
   publisher={IEEE}
 }

@article{VOC,
  title={The pascal visual object classes (voc) challenge},
  author={Everingham, Mark and Van Gool, Luc and Williams, Christopher KI and Winn, John and Zisserman, Andrew},
  journal={International Journal of Computer Vision},
  volume={88},
  number={2},
  pages={303--338},
  year={2010},
  publisher={Springer}
}

@article{adapter,
  title={Clip-adapter: Better vision-language models with feature adapters},
  author={Gao, Peng and Geng, Shijie and Zhang, Renrui and Ma, Teli and Fang, Rongyao and Zhang, Yongfeng and Li, Hongsheng and Qiao, Yu},
  journal={International Journal of Computer Vision},
  volume={132},
  number={2},
  pages={581--595},
  year={2024},
  publisher={Springer}
}

@article{Exdark,
  title = {Getting to Know Low-light Images with The Exclusively Dark Dataset},
  author = {Loh, Yuen Peng and Chan, Chee Seng},
  journal = {Computer Vision and Image Understanding},
  volume = {178},
  pages = {30-42},
  year = {2019},
  doi = {https://doi.org/10.1016/j.cviu.2018.10.010}
}

@article{trashcan,
  title={TrashCan 1.0 An Instance-Segmentation Labeled Dataset of Trash Observations},
  author={Hong, Jungseok and Fulton, Michael S and Sattar, Junaed},
  year={2020}
}

@article{DRYOLO,
  title={Dehazing \& Reasoning YOLO: Prior knowledge-guided network for object detection in foggy weather},
  author={Zhong, Fujin and Shen, Wenxin and Yu, Hong and Wang, Guoyin and Hu, Jun},
  journal={Pattern Recognition},
  pages={110756},
  year={2024},
  publisher={Elsevier}
}

@article{boostingRCNN,
  title={Boosting R-CNN: Reweighting R-CNN samples by RPN’s error for underwater object detection},
  author={Song, Pinhao and Li, Pengteng and Dai, Linhui and Wang, Tao and Chen, Zhan},
  journal={Neurocomputing},
  volume={530},
  pages={150--164},
  year={2023},
  publisher={Elsevier}
}

@inproceedings{dark-ruas,
title = {Retinex-inspired Unrolling with Cooperative Prior Architecture Search for Low-light Image Enhancement},
author = {Risheng, Liu and Long, Ma and Jiaao, Zhang and Xin, Fan and Zhongxuan, Luo},
booktitle = {Proceedings of the IEEE Conference on Computer Vision and Pattern Recognition},
year = {2021}
}

@inproceedings{dark-ZeroDCE,
 author = {Guo, Chunle Guo and Li, Chongyi and Guo, Jichang and Loy, Chen Change and Hou, Junhui and Kwong, Sam and Cong, Runmin},
 title = {Zero-reference deep curve estimation for low-light image enhancement},
 booktitle = {Proceedings of the IEEE conference on computer vision and pattern recognition (CVPR)},
 pages    = {1780-1789},
 month = {June},
 year = {2020}
}

@inproceedings{dark-UHDFour,
  title={EmbeddingFourier for Ultra-High-Definition
Low-Light Image Enhancement},
  author={Chongyi Li and Chun-Le Guo and  Man Zhou  and Zhexin Liang and  Shangchen Zhou and Ruicheng Feng and Chen Change Loy},
  booktitle={ICLR},
  year={2023}
}

@article{DCP,
  title={Single image haze removal using dark channel prior},
  author={He, Kaiming and Sun, Jian and Tang, Xiaoou},
  journal={IEEE transactions on pattern analysis and machine intelligence},
  volume={33},
  number={12},
  pages={2341--2353},
  year={2010},
  publisher={IEEE}
}

@inproceedings{PICLIP,
  title={Rethinking Prior Information Generation with CLIP for Few-Shot Segmentation},
  author={Wang, Jin and Zhang, Bingfeng and Pang, Jian and Chen, Honglong and Liu, Weifeng},
  booktitle={Proceedings of the IEEE/CVF Conference on Computer Vision and Pattern Recognition},
  pages={3941--3951},
  year={2024}
}

@article{VTCLIP,
  title={Visual and Textual Prior Guided Mask Assemble for Few-Shot Segmentation and Beyond},
  author={Chen, Shuai and Meng, Fanman and Zhang, Runtong and Qiu, Heqian and Li, Hongliang and Wu, Qingbo and Xu, Linfeng},
  journal={IEEE Transactions on Multimedia},
  year={2024},
  publisher={IEEE}
}

@inproceedings{RW1-AAAI,
  title={Simple image-level classification improves open-vocabulary object detection},
  author={Fang, Ruohuan and Pang, Guansong and Bai, Xiao},
  booktitle={Proceedings of the AAAI Conference on Artificial Intelligence},
  volume={38},
  number={2},
  pages={1716--1725},
  year={2024}
}

@inproceedings{RW2-CVPR,
  title={Sed: A simple encoder-decoder for open-vocabulary semantic segmentation},
  author={Xie, Bin and Cao, Jiale and Xie, Jin and Khan, Fahad Shahbaz and Pang, Yanwei},
  booktitle={Proceedings of the IEEE/CVF Conference on Computer Vision and Pattern Recognition},
  pages={3426--3436},
  year={2024}
}

@inproceedings{Intro-1,
  title={Detrs beat yolos on real-time object detection},
  author={Zhao, Yian and Lv, Wenyu and Xu, Shangliang and Wei, Jinman and Wang, Guanzhong and Dang, Qingqing and Liu, Yi and Chen, Jie},
  booktitle={Proceedings of the IEEE/CVF Conference on Computer Vision and Pattern Recognition},
  pages={16965--16974},
  year={2024}
}

@inproceedings{Intro-2,
  title={Semi-supervised Open-World Object Detection},
  author={Mullappilly, Sahal Shaji and Gehlot, Abhishek Singh and Anwer, Rao Muhammad and Khan, Fahad Shahbaz and Cholakkal, Hisham},
  booktitle={Proceedings of the AAAI Conference on Artificial Intelligence},
  volume={38},
  number={5},
  pages={4305--4314},
  year={2024}
}

@inproceedings{Intro-3,
  title={Fine-Grained Prototypes Distillation for Few-Shot Object Detection},
  author={Wang, Zichen and Yang, Bo and Yue, Haonan and Ma, Zhenghao},
  booktitle={Proceedings of the AAAI Conference on Artificial Intelligence},
  volume={38},
  number={6},
  pages={5859--5866},
  year={2024}
}

@ARTICLE{Intro-4,
  author={Quan, Yu and Zhang, Dong and Tang, Jinhui},
  journal={IEEE Transactions on Circuits and Systems for Video Technology}, 
  title={Generalized Concordant Vision Transformer with Masked Image Tokens for Object Detection}, 
  year={2025},
  volume={},
  number={},
  pages={1-1},
  doi={10.1109/TCSVT.2025.3570504}}

@ARTICLE{Intro-5,
  author={Feng, Juexiao and Yang, Yuhong and Lyu, Mengyao and Hao, Tianxiang and Huang, Yi-Jie and Xie, Yanchun and Li, Yaqian and Han, Jungong and Xiang, Liuyu and Ding, Guiguang},
  journal={IEEE Transactions on Circuits and Systems for Video Technology}, 
  title={Towards Realistic Hierarchical Object Detection: Problem, Benchmark and Solution}, 
  year={2025},
  volume={},
  number={},
  pages={1-1},
  doi={10.1109/TCSVT.2025.3552596}}

@ARTICLE{TCSVT-intro6,
  author={Ma, Zeyu and Zheng, Ziqiang and Wei, Jiwei and Yang, Yang and Shen, Heng Tao},
  journal={IEEE Transactions on Circuits and Systems for Video Technology}, 
  title={Instance-Dictionary Learning for Open-World Object Detection in Autonomous Driving Scenarios}, 
  year={2024},
  volume={34},
  number={5},
  pages={3395-3408},
  doi={10.1109/TCSVT.2023.3322465}}

@ARTICLE{TCSVT-intro7,
  author={Li, Huafeng and Yan, Shuanglin and Yu, Zhengtao and Tao, Dapeng},
  journal={IEEE Transactions on Circuits and Systems for Video Technology}, 
  title={Attribute-Identity Embedding and Self-Supervised Learning for Scalable Person Re-Identification}, 
  year={2020},
  volume={30},
  number={10},
  pages={3472-3485},
  doi={10.1109/TCSVT.2019.2952550}}

@ARTICLE{TCSVT-intro8,
  author={Dong, Neng and Zhang, Liyan and Yan, Shuanglin and Tang, Hao and Tang, Jinhui},
  journal={IEEE Transactions on Circuits and Systems for Video Technology}, 
  title={Erasing, Transforming, and Noising Defense Network for Occluded Person Re-Identification}, 
  year={2024},
  volume={34},
  number={6},
  pages={4458-4472},
  doi={10.1109/TCSVT.2023.3339167}}

@ARTICLE{TCSVT-intro9,
  author={Sun, Dengdi and Cheng, Leilei and Chen, Song and Li, Chenglong and Xiao, Yun and Luo, Bin},
  journal={IEEE Transactions on Circuits and Systems for Video Technology}, 
  title={UAV-Ground Visual Tracking: A Unified Dataset and Collaborative Learning Approach}, 
  year={2024},
  volume={34},
  number={5},
  pages={3619-3632},
  doi={10.1109/TCSVT.2023.3316990}}

@article{advancing-dehazy,
  title={Advancing Real-World Image Dehazing: Perspective, Modules, and Training},
  author={Feng, Yuxin and Ma, Long and Meng, Xiaozhe and Zhou, Fan and Liu, Risheng and Su, Zhuo},
  journal={IEEE Transactions on Pattern Analysis and Machine Intelligence},
  year={2024},
  publisher={IEEE}
}

@article{HCLR,
  title={HCLR-net: Hybrid contrastive learning regularization with locally randomized perturbation for underwater image enhancement},
  author={Zhou, Jingchun and Sun, Jiaming and Li, Chongyi and Jiang, Qiuping and Zhou, Man and Lam, Kin-Man and Zhang, Weishi and Fu, Xianping},
  journal={International Journal of Computer Vision},
  pages={1--25},
  year={2024},
  publisher={Springer}
}

@article{UFormer,
  title={U-shape transformer for underwater image enhancement},
  author={Peng, Lintao and Zhu, Chunli and Bian, Liheng},
  journal={IEEE Transactions on Image Processing},
  volume={32},
  pages={3066--3079},
  year={2023},
  publisher={IEEE}
}

@article{Relu,
  title={Rectified Linear Units Improve Restricted Boltzmann Machines},
  author={Nair, Vinod and Hinton, Geoffrey E.},
  journal={Proceedings of the 27th International Conference on Machine Learning (ICML-10)},
  pages={807--814},
  year={2010}
}

@inproceedings{deformable-detr,
  title={Deformable DETR: Deformable Transformers for End-to-End Object Detection},
  author={Zhu, Xizhou and Su, Weijie and Lu, Lewei and Li, Bin and Wang, Xiaogang and Dai, Jifeng},
  booktitle={International Conference on Learning Representations}
}

@inproceedings{rtdetr,
  title={Detrs beat yolos on real-time object detection},
  author={Zhao, Yian and Lv, Wenyu and Xu, Shangliang and Wei, Jinman and Wang, Guanzhong and Dang, Qingqing and Liu, Yi and Chen, Jie},
  booktitle={Proceedings of the IEEE/CVF conference on computer vision and pattern recognition},
  pages={16965--16974},
  year={2024}
}

@ARTICLE{TCSVT-CFMW,
  author={Li, Haoyuan and Hu, Qi and Zhou, Binjia and Yao, You and Lin, Jiacheng and Yang, Kailun and Chen1, Peng},
  journal={IEEE Transactions on Circuits and Systems for Video Technology}, 
  title={CFMW: Cross-modality Fusion Mamba for Robust Object Detection under Adverse Weather}, 
  year={2025},
  volume={},
  number={},
  pages={1-1},
  doi={10.1109/TCSVT.2025.3587918}}

@ARTICLE{TCSVT-VLM,
  author={Xu, Shilin and Li, Xiangtai and Wu, Size and Zhang, Wenwei and Tong, Yunhai and Change Loy, Chen},
  journal={IEEE Transactions on Circuits and Systems for Video Technology}, 
  title={DST-Det: Open-Vocabulary Object Detection via Dynamic Self-Training}, 
  year={2025},
  volume={35},
  number={5},
  pages={5037-5050},
  doi={10.1109/TCSVT.2024.3520734}}

@ARTICLE{TCSVT-VLM2,
  author={Ma, Chengcheng and Liu, Yang and Deng, Jiankang and Xie, Lingxi and Dong, Weiming and Xu, Changsheng},
  journal={IEEE Transactions on Circuits and Systems for Video Technology}, 
  title={Understanding and Mitigating Overfitting in Prompt Tuning for Vision-Language Models}, 
  year={2023},
  volume={33},
  number={9},
  pages={4616-4629},
  doi={10.1109/TCSVT.2023.3245584}}
\vspace{-10mm}
\begin{IEEEbiography}[{\includegraphics[width=1in,height=1.25in,clip,keepaspectratio]{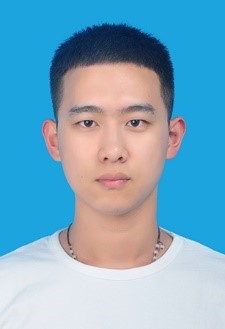}}]
{Jian Pang} is currently a Ph.D. candidate at the College of Control Science and Engineering, China University of Petroleum (East China), Qingdao, PR China. He received the B.S. degree in electronic information engineering from Qiqihar University, Qiqihar, PR China, in 2018, and the M.E. degree in electronic and communication engineering from Kunming University of Science and Technology, Kunming, PR China, in 2021. His main research interests include image enhancement, image re-identification, and object detection.
\end{IEEEbiography}
\vspace{-10mm}
\begin{IEEEbiography}[{\includegraphics[width=1in,height=1.25in,clip,keepaspectratio]{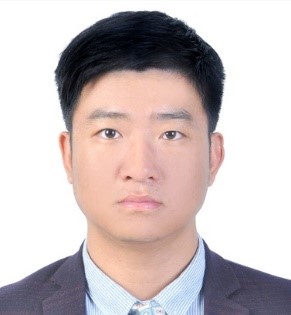}}]
{Bingfeng Zhang} received the B.S. degree in electronic information engineering from China University of Petroleum (East China), Qingdao, PR China, in 2015, the M.E. degree in systems, control and signal processing from University of Southampton, Southampton, U.K., in 2016. And the Ph.D degree from the University of Liverpool, Liverpool, U.K. His current research interest is computer vision, including semantic segmentation with limited annotation, salient object detection, and video object segmentation.
\end{IEEEbiography}
\vspace{-10mm}
\begin{IEEEbiography}[{\includegraphics[width=1in,height=1.25in,clip,keepaspectratio]{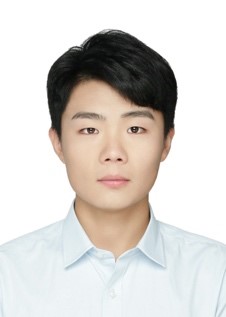}}]
{Jin Wang} received the B.S. degree in automation from Qingdao University of Science and Technology, Qingdao, PR China, in 2021. He is currently pursuing the Ph.D. degree with China University of Petroleum (East China), Qingdao, PR China. His research interests include computer vision, deep learning, and pattern recognition, especially semantic segmentation.
\end{IEEEbiography}
\vspace{-10mm}
\begin{IEEEbiography}[{\includegraphics[width=1in,height=1.25in,clip,keepaspectratio]{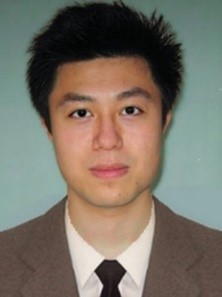}}]
{Baodi Liu} received the Ph.D. degree in Electronic Engineering from Tsinghua University. Currently, he is an assistant professor in the College of Information and Control Engineering, China University of Petroleum, China. His research interests include computer vision and machine learning.
\end{IEEEbiography}
\vspace{-10mm}
\begin{IEEEbiography}[{\includegraphics[width=1in,height=1.25in,clip,keepaspectratio]{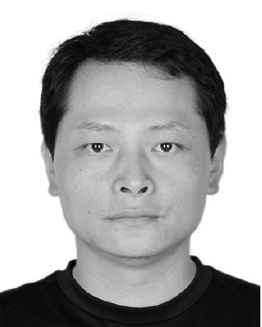}}]
{Dapeng Tao} received the B.E. degree from Northwestern Polytechnical University, Xi’an, China, in 1999, and the Ph.D. degree from the South China University of Technology, Guangzhou, China, in 2014. He is currently a Professor with the School of Information Science and Engineering, Yunnan University, Kunming, China. He has authored and coauthored more than 50 scientific articles. He has served more than ten international journals, including the IEEE Transactions on Image Processing, the IEEE Transactions on Neural Networks and Learning Systems, the IEEE Transactions on Cybernetics, the IEEE Transactions on Multimedia, the IEEE Transactions on Circuits and Systems for Video Technology, Pattern Recognition, and Information Sciences. His research interests include machine learning, computer vision, and robotics.
\end{IEEEbiography}
\vspace{-10mm}
\begin{IEEEbiography}[{\includegraphics[width=1in,height=1.25in,clip,keepaspectratio]{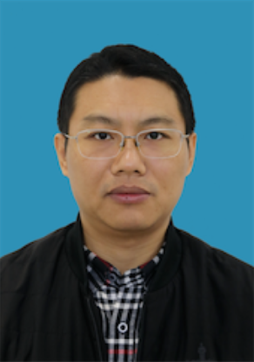}}]
{Weifeng Liu} is currently a Professor with the College of Control Science and Engineering, China University of Petroleum (East China), China. He received the double B.S. degree in automation and business administration and the Ph.D. degree in pattern recognition and intelligent systems from the University of Science and Technology of China, Hefei, China, in 2002 and 2007, respectively. His current research interests include pattern recognition and machine learning. He has authored or co-authored a dozen papers in top journals and prestigious conferences including 10 ESI Highly Cited Papers and 3 ESI Hot Papers. Dr. Weifeng Liu serves as associate editor for Neural Processing Letter, co-chair for IEEE SMC technical committee on cognitive computing, and guest editor of special issue for Signal Processing, IET Computer Vision, Neurocomputing, and Remote Sensing. He also serves dozens of journals and conferences.
\end{IEEEbiography}

\end{document}